\documentclass[runningheads]{llncs}
\usepackage[T1]{fontenc}
\usepackage{graphicx,verbatim}
\usepackage{amsmath,amssymb,booktabs,longtable}
\usepackage{multirow} 
\usepackage[table]{xcolor} 
\usepackage{pgfplots}
\pgfplotsset{compat=1.18}
\usepackage{tikz}
\usetikzlibrary{shapes,arrows,positioning,shadows,backgrounds,fit}
\usepackage{placeins}
\begin{document}

\title{From Retinal Evidence to Safe Decisions: RETINA-SAFE and ECRT for Hallucination Risk Triage in Medical LLMs}
\titlerunning{RETINA-SAFE and ECRT for Medical LLM Risk Triage}

\author{Zhe Yu\inst{1,2}\thanks{These authors contributed equally to this work.} \and 
        Wenpeng Xing\inst{1,3}\protect\footnotemark[1] \and 
        Meng Han\inst{1,3,4}\thanks{Corresponding author.}}
\authorrunning{Z. Yu et al.}

\institute{
Binjiang Institute of Zhejiang University, Hangzhou, China \and
Communication University of Zhejiang, Hangzhou, China \and           
           Zhejiang University, Hangzhou, China \and
           Gentel.io\\
           \email{zyu@zju-if.com, \{wpxing, mhan\}@zju.edu.cn}}

\maketitle
\begin{abstract}
Hallucinations in medical large language models (LLMs) remain a safety-critical issue, particularly when available evidence is insufficient or conflicting. We study this problem in diabetic retinopathy (DR) decision settings and introduce RETINA-SAFE, an evidence-grounded benchmark aligned with retinal grading records, comprising 12,522 samples. RETINA-SAFE is organized into three evidence-relation tasks: E-Align (evidence-consistent), E-Conflict (evidence-conflicting), and E-Gap (evidence-insufficient). We further propose ECRT (Evidence-Conditioned Risk Triage), a two-stage white-box detection framework: Stage 1 performs Safe/Unsafe risk triage, and Stage 2 refines unsafe cases into contradiction-driven versus evidence-gap risks. ECRT leverages internal representation and logit shifts under CTX/NOCTX conditions, with class-balanced training for robust learning. Under evidence-grouped (not patient-disjoint) splits across multiple backbones, ECRT provides strong Stage-1 risk triage and explicit subtype attribution, improves Stage-1 balanced accuracy by +0.15 to +0.19 over external uncertainty and self-consistency baselines and by +0.02 to +0.07 over the strongest adapted supervised baseline, and consistently exceeds a single-stage white-box ablation on Stage-1 balanced accuracy. These findings support white-box internal signals grounded in retinal evidence as a practical route to interpretable medical LLM risk triage.

\end{abstract}

\section{Introduction}
Medical large language models (LLMs) are being adopted for clinical decision support, but reliability remains a concern~\cite{pandit2025medhallu}. In diabetic retinopathy (DR), unsafe recommendations can occur when evidence is missing or conflicting under grading workflows~\cite{wilkinson2003drscale}. Most evaluations focus on answer correctness, while evidence-\hspace{0pt}conditioned safety modes like contradiction and insufficiency are less explored. 

This paper treats hallucination risk as an evidence-\hspace{0pt}conditioned triage problem. We define three settings: evidence-consistent (E-Align), evidence-conflicting (E-Conflict), and evidence-insufficient (E-Gap). Our contributions include: (1) \textbf{RETINA-SAFE}, a DR-grounded benchmark with 12,522 samples aligned with retinal grading; (2) \textbf{ECRT}, a two-stage white-box triage framework using internal CTX/NOCTX representation shifts; and (3) a systematic multi-backbone analysis under a leakage-aware protocol. ECRT provides strong risk triage, outperforming generic uncertainty baselines. Related work has explored medical LLM reliability~\cite{pal2023medhalt,hardy2025rextrust} and white-box diagnostics~\cite{chen2024inside,azaria2023internal,zhang2023focus,manakul2023selfcheckgpt}, but clinically interpretable risk triage with explicit evidence relations remains underexplored.

\section{RETINA-SAFE: An Evidence-Grounded Benchmark}
\paragraph{Clinical Provenance and Evidence Schema.}
RETINA-SAFE is constructed from retinal grading records and aligned evidence metadata, yielding 12,522 samples. Dataset construction, evidence canonicalization, and task design were conducted with clinical input and review from hospital-affiliated co-authors and clinical collaborators. Each sample contains (1) a clinical-style question, (2) four options, (3) a gold answer, and (4) evidence text derived from DR grading signals~\cite{wilkinson2003drscale,etdrs1991early,ada2024standards}. It uses \emph{image-derived textual evidence} as a structured intermediate representation. The benchmark evaluates evidence-grounded reliability: can the system decide if a recommendation is trustworthy, contradicted, or unsupported?

\begin{figure}[!t]
\centering
\resizebox{0.85\linewidth}{!}{%
\begin{tikzpicture}[
    node distance=1.0cm and 1.2cm,
    font=\sffamily\small,
    >=latex,
    recordBox/.style={
        rectangle, rounded corners=4pt, thick,
        draw=gray!60!black, top color=gray!5, bottom color=gray!10,
        minimum width=3.0cm, minimum height=2.4em,
        drop shadow={opacity=0.1}, align=center
    },
    processBox/.style={
        rectangle, rounded corners=4pt, thick,
        draw=teal!60!black, top color=teal!5, bottom color=teal!10,
        minimum width=3.0cm, minimum height=2.4em,
        drop shadow={opacity=0.1}, align=center
    },
    schemaBox/.style={
        rectangle, rounded corners=4pt, thick,
        draw=blue!60!black, top color=blue!5, bottom color=blue!10,
        minimum width=3.0cm, minimum height=2.4em,
        drop shadow={opacity=0.1}, align=center
    },
    alignBox/.style={
        rectangle, rounded corners=4pt, thick,
        draw=green!60!black, top color=green!5, bottom color=green!10,
        minimum width=3.5cm, minimum height=3.0em,
        drop shadow={opacity=0.1}, align=center
    },
    conflictBox/.style={
        rectangle, rounded corners=4pt, thick,
        draw=red!60!black, top color=red!5, bottom color=red!10,
        minimum width=3.5cm, minimum height=3.0em,
        drop shadow={opacity=0.1}, align=center
    },
    gapBox/.style={
        rectangle, rounded corners=4pt, thick,
        draw=orange!60!black, top color=orange!5, bottom color=orange!10,
        minimum width=3.5cm, minimum height=3.0em,
        drop shadow={opacity=0.1}, align=center
    },
    groupLabel/.style={font=\bfseries\footnotesize, color=gray!60!black, anchor=south west, yshift=2pt},
    moduleBox/.style={rectangle, draw=gray!40, dashed, fill=gray!2, rounded corners=8pt, inner sep=10pt, line width=1pt}
]

    \node[recordBox] (raw) {Raw Retinal Records\\(Clinical Prose \& Images)};
    \node[processBox, right=1.0cm of raw] (rules) {Rule-Based Extraction\\\& Canonicalization};
    \node[schemaBox, right=1.0cm of rules] (schema) {Structured Schema\\$(q, o, c, e)$};
    
    \draw[->, thick, color=gray!80!black] (raw.east) -- (rules.west);
    \draw[->, thick, color=teal!80!black] (rules.east) -- (schema.west);

    \begin{scope}[on background layer]
        \node[moduleBox, fit=(raw) (rules) (schema)] (constructModule) {};
        \node[groupLabel] at (constructModule.north west) {Clinical Provenance Data Flow};
    \end{scope}

    \node[schemaBox, below=1.8cm of rules] (input) {Medical Prompt + Evidence\\(Test Sample)};
    
    \node[conflictBox, below=1.5cm of input, xshift=-0.18cm] (econflict) {\textbf{E-Conflict}\\Evidence contradicts\\false premise in prompt};
    \node[alignBox, left=0.6cm of econflict, xshift=-0.08cm] (ealign) {\textbf{E-Align}\\Correct answer is\\directly supported};
    \node[gapBox, right=0.6cm of econflict, xshift=0.18cm] (egap) {\textbf{E-Gap}\\Evidence is insufficient\\(requires clinical deferment)};
    
    \draw[->, thick, color=blue!60!black, rounded corners=6pt] (schema.south) |- (input.east);
    
    \draw[->, thick, color=green!60!black] (input.south) -- node[above left, font=\footnotesize, xshift=1em, yshift=0em] {Consistent} (ealign.north);
    \draw[->, thick, color=red!60!black] (input.south) -- node[right, font=\footnotesize] {Contradiction} (econflict.north);
    \draw[->, thick, color=orange!60!black] (input.south) -- node[above right, font=\footnotesize, xshift=0.4em, yshift=-0.5em] {Insufficient} (egap.north);

    \begin{scope}[on background layer]
        \node[moduleBox, fit=(input) (ealign) (egap) (econflict)] (taxModule) {};
        \node[groupLabel] at (taxModule.north west) {Three-Way Evidence-Relation Taxonomy};
    \end{scope}

\end{tikzpicture}%
}
\caption{\textbf{RETINA-SAFE Benchmark Construction and Taxonomy.} Driven by raw clinical retinal records, structural attributes are extracted to formulate cases $(q,o,c,e)$. The evaluation follows a safety-critical taxonomy: answers are either directly supported (E-Align), structurally conflicting with the prompt (E-Conflict), or lacking sufficient evidence to decide (E-Gap).}
\label{fig:retinasafe_taxonomy}
\end{figure}
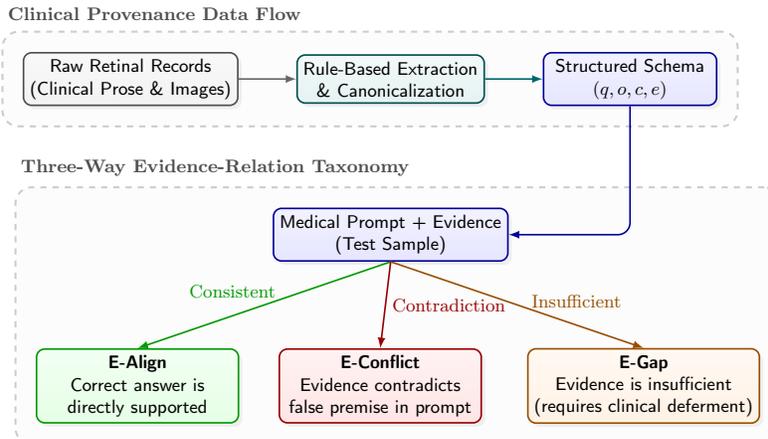

\paragraph{Taxonomy and Construction.}
RETINA-SAFE defines three categories (Fig.~\ref{fig:retinasafe_taxonomy}): (1) \textbf{E-Align}: directly supported; (2) \textbf{E-Conflict}: prompt contains misleading cues rejected by evidence; (3) \textbf{E-Gap}: evidence is insufficient, requiring clinical deferment. Items are rule-constructed from structured retinal grading descriptors. E-Gap captures intentional evidence insufficiency. Table~\ref{tab:retina_stats} summarizes the dataset (including class ratios), and Table~\ref{tab:retina_benchmark_baselines} shows that standard LLMs (e.g., Llama 3-8B) struggle, primarily failing to abstain on E-Gap cases.

\begin{table}[tb]
\caption{RETINA-SAFE benchmark statistics (computed from \texttt{retina\_safe.jsonl}; lengths use whitespace token counts for question and evidence text).}
\label{tab:retina_stats}
\centering
\scriptsize
\begin{tabular}{@{}lcccc@{}}
\toprule
Class / Split & Samples & Ratio & Avg Q Len & Avg Ev Len \\
\midrule
E-Align & 1,149 & 0.092 & 34.15 & 3.00 \\
E-Conflict & 5,107 & 0.408 & 47.34 & 3.00 \\
E-Gap & 6,266 & 0.500 & 38.67 & 3.00 \\
\midrule
All & 12,522 & 1.000 & 41.79 & 3.00 \\
\bottomrule
\end{tabular}
\end{table}

\begin{table}[tb]
\caption{RETINA-SAFE benchmark difficulty baselines (black-box MCQA with evidence). Columns report per-task accuracy and macro task accuracy.}
\label{tab:retina_benchmark_baselines}
\centering
\scriptsize
\begin{tabular}{@{}lcccc@{}}
\toprule
Baseline & E-Align & E-Conflict & E-Gap & Macro-Acc \\
\midrule
Random-choice (4-way; lower bound) & 0.2500 & 0.2500 & 0.2500 & 0.2500 \\
Llama 3-8B (black-box QA) & 0.4412 & 0.2215 & 0.1241 & 0.2623 \\
Llama 2-7B (black-box QA) & 0.3842 & 0.1412 & 0.0815 & 0.2023 \\
Llama 2-13B (black-box QA) & 0.5124 & 0.1642 & 0.0912 & 0.2559 \\
Qwen2.5-7B (black-box QA) & 0.5891 & 0.8812 & 0.2114 & 0.5606 \\
\bottomrule
\end{tabular}
\end{table}

\begin{figure}[tb]
\centering
\begin{minipage}[t]{0.62\linewidth}
\centering
\setlength{\tabcolsep}{0pt}
\renewcommand{\arraystretch}{1.0}
\begin{tabular}{@{}c@{\hspace{4pt}}c@{\hspace{4pt}}c@{}}
\includegraphics[width=0.30\linewidth]{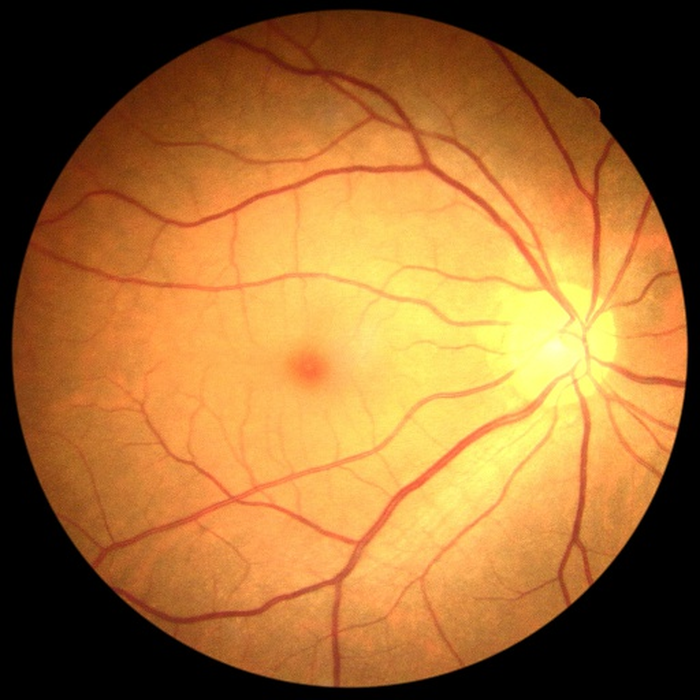} &
\includegraphics[width=0.30\linewidth]{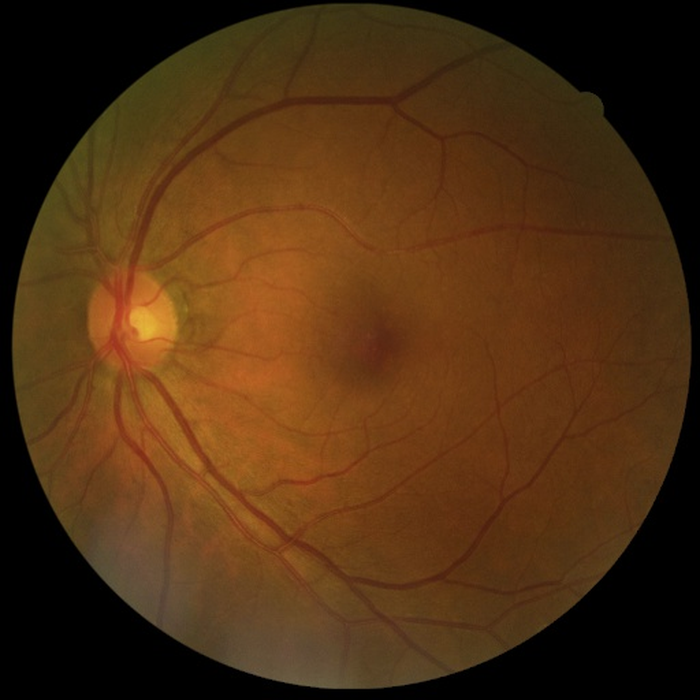} &
\includegraphics[width=0.30\linewidth]{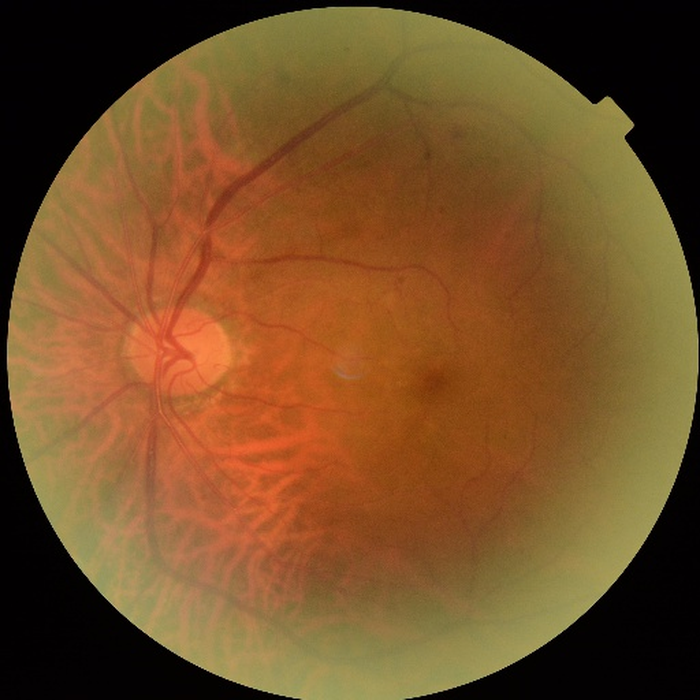} \\[-4pt]
\parbox[t]{0.30\linewidth}{\centering\scriptsize \textbf{(a)} E-Gap\\[-1pt] no DR} &
\parbox[t]{0.30\linewidth}{\centering\scriptsize \textbf{(b)} E-Conflict\\[-1pt] mild NPDR} &
\parbox[t]{0.30\linewidth}{\centering\scriptsize \textbf{(c)} E-Conflict\\[-1pt] moderate NPDR} \\[8pt]
\includegraphics[width=0.30\linewidth]{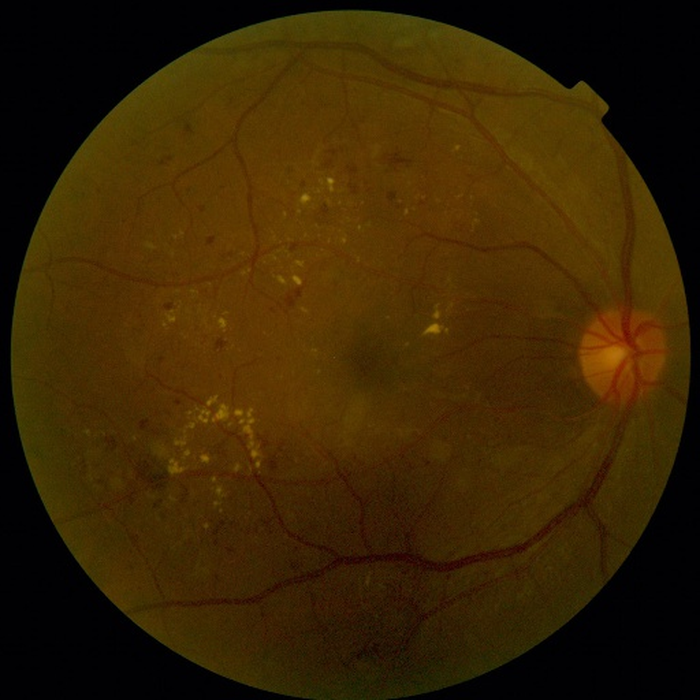} &
\includegraphics[width=0.30\linewidth]{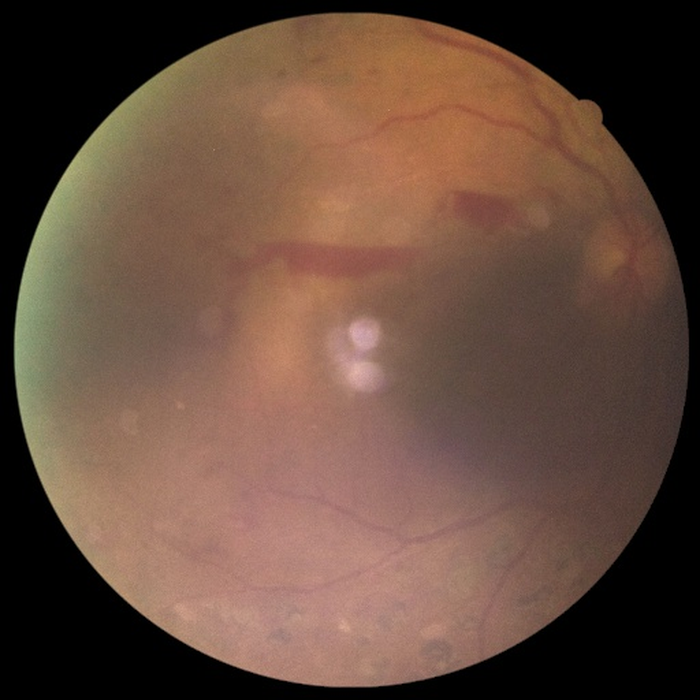} &
\includegraphics[width=0.30\linewidth]{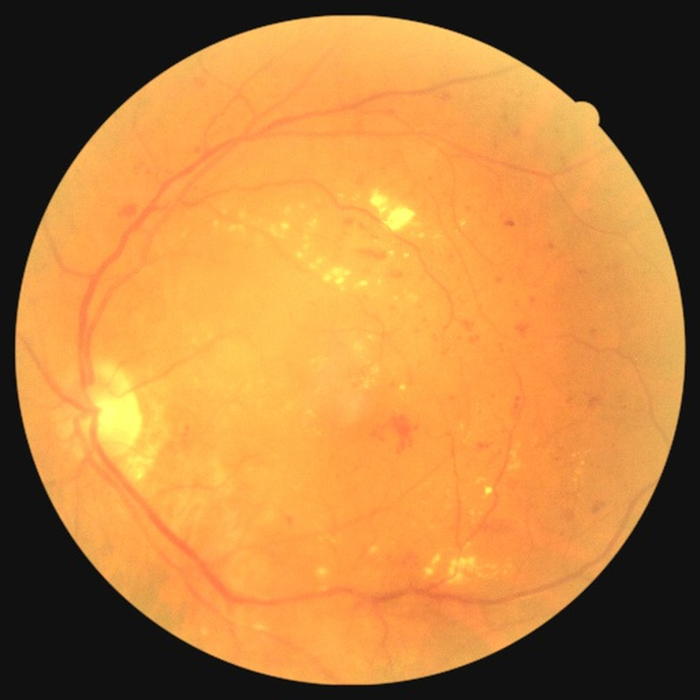} \\[-4pt]
\parbox[t]{0.30\linewidth}{\centering\scriptsize \textbf{(d)} E-Align\\[-1pt] severe NPDR} &
\parbox[t]{0.30\linewidth}{\centering\scriptsize \textbf{(e)} E-Align\\[-1pt] proliferative DR} &
\parbox[t]{0.30\linewidth}{\centering\scriptsize \textbf{(f)} E-Align\\[-1pt] severe NPDR}
\end{tabular}
\end{minipage}\hfill
\raisebox{30mm}{%
\begin{minipage}[t]{0.35\linewidth}
\vspace{0pt}
\scriptsize
\setlength{\tabcolsep}{2pt}
\renewcommand{\arraystretch}{1.03}
\textbf{RETINA-SAFE record schema}\par
\vspace{2pt}
\begin{tabular}{@{}ll@{}}
\toprule
Field & Meaning \\
\midrule
$q$ & clinical question \\
$o$ & 4-way options \\
$c$ & context \\
$e$ & structured retinal evidence \\
\bottomrule
\end{tabular}

\vspace{5pt}
\textbf{Task semantics}\par
\vspace{2pt}
\begin{tabular}{@{}p{0.26\linewidth}p{0.70\linewidth}@{}}
\toprule
Task & Decision logic \\
\midrule
E-Align & Evidence directly supports the answer (safe proceed). \\
E-Conflict & Evidence rejects a false prompt premise (contradiction risk). \\
E-Gap & Evidence is insufficient; defer or request more information. \\
\bottomrule
\end{tabular}

\end{minipage}%
}
\caption{\textbf{RETINA-SAFE benchmark exemplars (images + task semantics).} Left: representative retinal images linked to evidence records across E-Align / E-Conflict / E-Gap. Right: the RETINA-SAFE textual record schema and task semantics, illustrating that labels are assigned from the evidence-\hspace{0pt}conditioned decision setting (question + structured evidence), not from image appearance alone; labels are defined by the question--evidence relation rather than image severity alone.}
\label{fig:retinasafe_examples_v1}
\end{figure}

\paragraph{Per-class difficulty pattern.}
Table~\ref{tab:retina_benchmark_baselines} reveals a clear asymmetry: E-Gap remains difficult across black-box baselines (e.g., 0.1241 for Llama 3-8B and 0.0912 for Llama 2-13B) despite being the largest class, and Qwen2.5-7B shows a large within-model gap between E-Conflict (0.8812) and E-Gap (0.2114). This indicates that insufficiency detection is not recovered by class frequency alone, and that contradiction handling versus deferment behavior are separable capabilities rather than a single overall QA strength factor. These patterns motivate explicit contradiction-vs-gap decomposition and gap-aware safety metrics. The short average evidence length in Table~\ref{tab:retina_stats} reflects canonicalized retinal descriptors, and Fig.~\ref{fig:retinasafe_examples_v1} provides representative linked images and task semantics.

\paragraph{Leakage-Aware Protocol.}
\label{sec:retina_protocol}
We define a strict evaluation protocol keyed by the evidence identifier field (\texttt{evidence\_}\hspace{0pt}\texttt{id\_}\hspace{0pt}\texttt{code}). RETINA-SAFE includes semantically related items that can share evidence templates. To reduce leakage, splits use grouped stratification (nested \texttt{GroupKFold}-style), threshold policies are selected on validation only, and final test performance is reported once with frozen thresholds. 

\section{The ECRT Framework}
\label{sec:ecrt_framework}

The ECRT framework treats hallucination detection as a supervised binary-or-ternary triage task. Given input $x=(q,o,c,e)$, ECRT predicts a risk label in $\{\text{safe},\text{unsafe-contradiction},\text{unsafe-gap}\}$.

\begin{figure}[t]
\centering
\includegraphics[width=\linewidth]{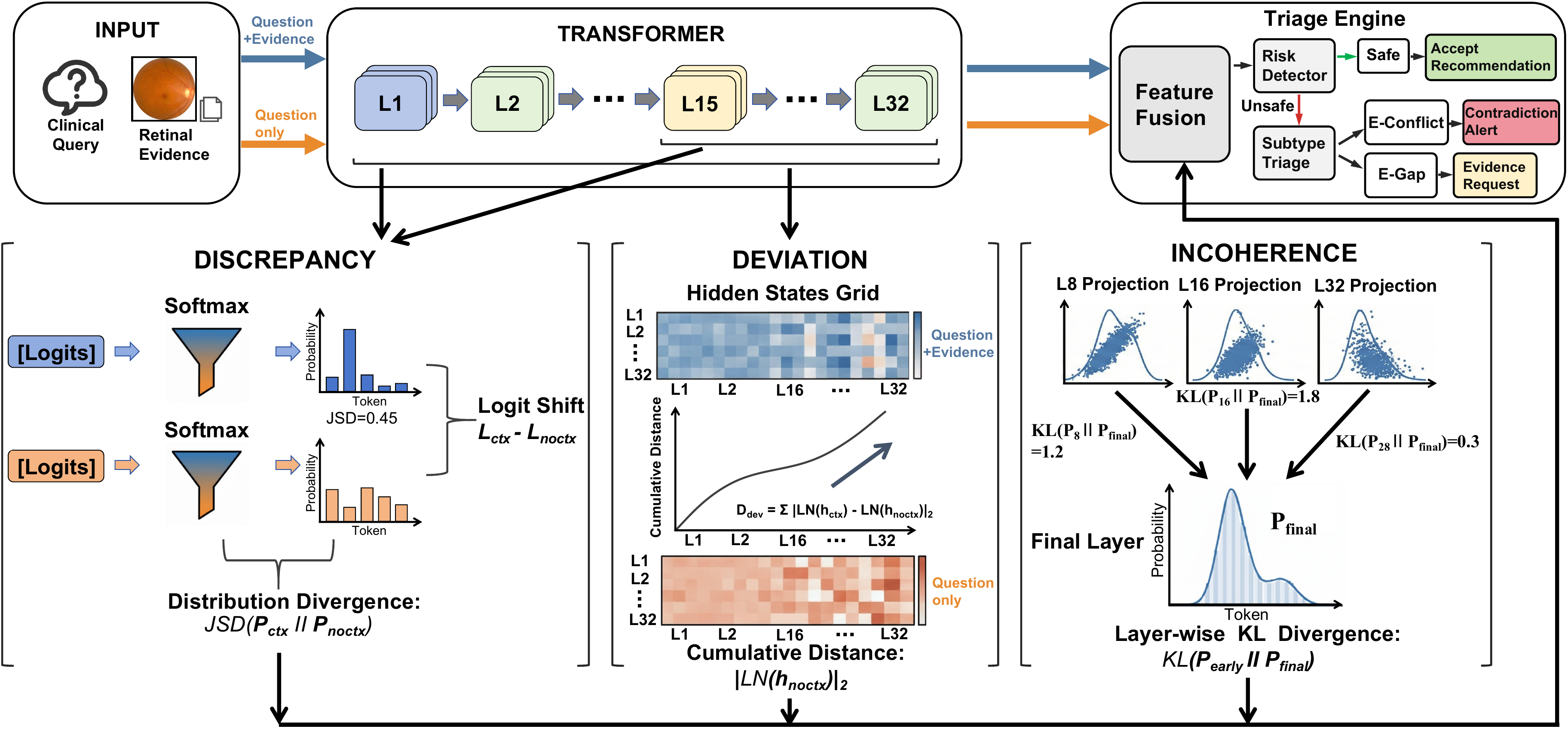}
\caption{\textbf{ECRT Framework Overview.} ECRT leverages internal representation and logit shifts under CTX/NOCTX conditions to extract three families of safety signals (Discrepancy, Deviation, Incoherence). These features train a two-stage triage engine that detects risks and attributes them to clinical contradiction or evidence-gap categories.}
\label{fig:ecrt_framework}
\end{figure}

\paragraph{Internal Signal Families.}
ECRT extracts three families of internal signals (Table~\ref{tab:ecrt_signals}) from the paired CTX/NOCTX passes.
(1) \textbf{Discrepancy} ($\Delta \mathrm{logits}$): captures the alignment shift at the final prediction layer $L$, where $\Delta z_t = z_{t,\mathrm{CTX}} - z_{t,\mathrm{NOCTX}}$. 
(2) \textbf{Deviation} ($\Delta \mathrm{traj}$): captures the internal trajectory shift across all layers $\ell \in [1, L]$, defined as $\Delta h_t^{\ell} = h_{t,\mathrm{CTX}}^{\ell} - h_{t,\mathrm{NOCTX}}^{\ell}$. 
(3) \textbf{Incoherence} ($\Delta \mathrm{inc}$): captures the semantic stability of the model across individual tokens $t$ and layers $\ell$. 

Current ECRT operates on structured, image-derived retinal descriptors in $e$ (e.g., grading findings extracted from retinal records), enabling controlled CTX/NOCTX perturbations and interpretable internal signal analysis. In this sense, image evidence directly influences the measured internal shifts through the structured evidence channel: adding or removing image-supported findings changes the model's latent trajectory, logits, and token-level consistency. This provides a clinically interpretable bridge from retinal evidence to white-box risk signals. Extending the same framework with end-to-end pixel-level multimodal features is a natural next step.

\begin{table}[t]
\caption{ECRT Internal Signal Families. Features are extracted via teacher-forced paired passes and aggregated into a pooled vector $\mathcal{P}(S)$.}
\label{tab:ecrt_signals}
\centering
\scriptsize
\begin{tabular}{@{}lllc@{}}
\toprule
Family & Notation & Level & Technical Metric \\
\midrule
Discrepancy & $\Delta \mathrm{logits}$ & Prediction & Logit output shift $\Delta z_t$ \\
Deviation & $\Delta \mathrm{traj}$ & Latent & Hidden-state shift $\Delta h_t^{\ell}$ \\
Incoherence & $\Delta \mathrm{inc}$ & Structural & Per-layer KL-Divergence \\
\bottomrule
\end{tabular}
\end{table}

\paragraph{Algorithm and Two-Stage Triage.}
ECRT decomposes the safety triage into two distinct stages. 

(1) \textbf{Stage 1}: estimate $p_{\mathrm{unsafe}}$, which indicates if the response is either a contradiction or an information gap. 
(2) \textbf{Stage 2}: for cases flagged as unsafe, estimate $p_{\mathrm{gap}\mid\mathrm{unsafe}}$ to distinguish between lack of evidence (Gap) and active falsification (Conflict). 
Final probabilities are composed as $p_{\mathrm{align}}=1-p_{\mathrm{unsafe}}$, $p_{\mathrm{contradict}}=p_{\mathrm{unsafe}}(1-p_{\mathrm{gap}\mid\mathrm{unsafe}})$, and $p_{\mathrm{gap}}=p_{\mathrm{unsafe}}p_{\mathrm{gap}\mid\mathrm{unsafe}}$.

\section{Experiments}
\label{sec:experiments}
\paragraph{Implementation Details.} Our experiments were conducted on a server equipped with six NVIDIA GeForce RTX 4090 GPUs (24GB VRAM). LLM backbones were loaded in 16-bit precision. The ECRT triage heads utilize XGBoost classifiers with 160 estimators, trained to optimize for the target-recall policy ($\tau=0.95$) on a held-out validation set. Feature extraction for CTX/NOCTX probes was performed with a batch size of 1, preserving hidden states across all transformer layers.

\paragraph{Setup.}
We evaluate Llama 3-8B, Llama 2-7B, Mistral-7B, Qwen2.5-7B, and Qwen3-8B on backbone-specific generated response corpora, using one fixed response per item and freezing responses for all detector comparisons. ECRT, the single-stage white-box ablation, and external baselines share the same responses and grouped split policy (seed13 for Table~\ref{tab:external_baselines_stage1}); CTX/NOCTX passes use teacher-forced alignment on generated tokens. For parity, all detector comparisons on a given backbone use the same frozen generations and split policy, so gains are not attributable to decoding variance. When white-box extraction is unavailable, comparisons are restricted to matched subsets. Table~\ref{tab:retina_benchmark_baselines} separately reports black-box MCQA difficulty. Endpoints are Stage-1 (Recall, Flag Rate, BA) and Stage-2 (Gap and Contradiction Recall), with validation-only threshold tuning.

\begin{figure}[t]
\centering
\begin{tikzpicture}
    \begin{axis}[
        ybar=2pt,
        bar width=10pt,
        width=0.95\linewidth,
        height=4.8cm,
        enlarge x limits=0.15,
        ymin=0.80, ymax=1.05,
        ytick={0.80, 0.85, 0.90, 0.95, 1.00},
        ylabel={Stage-1 BA},
        symbolic x coords={L3-8B, L2-7B, Mistral, Qwen2.5, Qwen3},
        xtick=data,
        nodes near coords,
        every node near coord/.append style={font=\tiny, rotate=90, anchor=west, xshift=2pt},
        legend style={at={(0.5,-0.25)}, anchor=north, legend columns=-1, draw=none},
        ymajorgrids=true,
        grid style=dashed,
        tick label style={font=\scriptsize},
        label style={font=\small}
    ]
    \addplot[fill=gray!30, draw=gray!80, thick] coordinates {
        (L3-8B, 0.8124) (L2-7B, 0.8926) (Mistral, 0.9085) (Qwen2.5, 0.9412) (Qwen3, 0.9324)
    };
    \addplot[fill=blue!40!white, draw=blue!80!black, thick] coordinates {
        (L3-8B, 0.8405) (L2-7B, 0.9229) (Mistral, 0.9310) (Qwen2.5, 0.9679) (Qwen3, 0.9640)
    };
    \legend{1-stage, ECRT (\textbf{Ours})}
    \end{axis}
\end{tikzpicture}
\caption{Comparison of primary clinical endpoint (Stage-1 BA) under a target-recall policy. ECRT consistently outperforms the single-stage ablation across all evaluated backbones.}
\label{fig:primary_bar}
\end{figure}
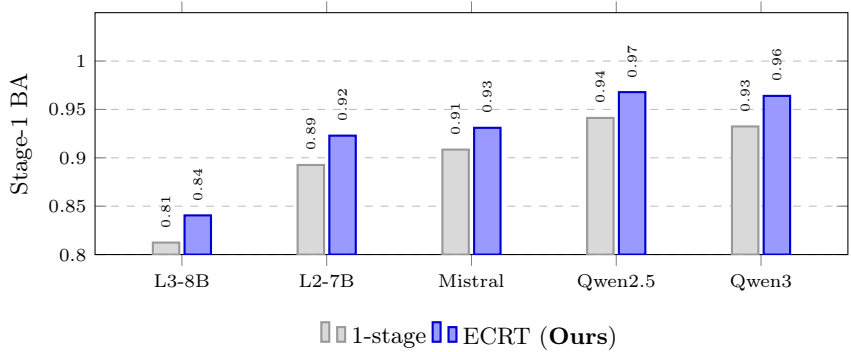

\paragraph{Triage Performance (Stage 1).}
Table~\ref{tab:primary_main} reports the primary clinical endpoint. Across backbones, ECRT consistently exceeds the single-stage baseline on Stage-1 BA, supporting the two-stage decomposition. Llama 3-8B is the hardest regime, while Qwen models show strong separability. Fig.~\ref{fig:primary_bar} visualizes the BA improvements. Table~\ref{tab:external_baselines_stage1} compares ECRT against external baselines under a high-recall policy. Relative to uncertainty and self-consistency baselines, ECRT improves Stage-1 BA by +0.15 to +0.19; relative to the strongest adapted supervised baseline, gains remain +0.02 to +0.07 while maintaining lower review burden.

\begin{table}[t]
\caption{Primary endpoint (Stage 1 Safe/Unsafe) under the strict group protocol using the target-recall policy. Values are mean $\pm$ std across seeds (Llama 3-8B: 5; others: 3).}
\label{tab:primary_main}
\centering
\scriptsize
\begin{tabular}{@{}llccc@{}}
\toprule
Backbone & Method & U-Recall & Flag Rate & S1 BA \\
\midrule
L3-8B & ECRT & $0.9505 \pm 0.0011$ & $0.8712 \pm 0.0049$ & $0.8405 \pm 0.0193$ \\
L3-8B & 1-stage & $0.9556 \pm 0.0026$ & $0.8767 \pm 0.0054$ & $0.8124 \pm 0.0201$ \\
\midrule
L2-7B & ECRT & $0.9430 \pm 0.0076$ & $0.8637 \pm 0.0151$ & $0.9229 \pm 0.0132$ \\
L2-7B & 1-stage & $0.9482 \pm 0.0099$ & $0.8878 \pm 0.0216$ & $0.8926 \pm 0.0117$ \\
\midrule
Mistral & ECRT & $0.9466 \pm 0.0024$ & $0.8636 \pm 0.0048$ & $0.9310 \pm 0.0024$ \\
Mistral & 1-stage & $0.9458 \pm 0.0024$ & $0.8802 \pm 0.0051$ & $0.9064 \pm 0.0021$ \\
\midrule
Qwen2.5-7B & ECRT & $0.9566 \pm 0.0024$ & $0.8665 \pm 0.0022$ & $0.9679 \pm 0.0037$ \\
Qwen2.5-7B & 1-stage & $0.9527 \pm 0.0016$ & $0.8623 \pm 0.0017$ & $0.9412 \pm 0.0018$ \\
\midrule
Qwen3-8B & ECRT & $0.9488 \pm 0.0048$ & $0.8595 \pm 0.0054$ & $0.9640 \pm 0.0014$ \\
Qwen3-8B & 1-stage & $0.9471 \pm 0.0008$ & $0.8580 \pm 0.0010$ & $0.9324 \pm 0.0021$ \\
\bottomrule
\end{tabular}
\end{table}

\noindent\textbf{Notation.} In Table~\ref{tab:primary_main}, \textit{1-stage} denotes a \textit{Single-stage white-box} baseline, i.e., the same white-box feature space and classifier family as ECRT without the two-stage decomposition.

\subsection{External Baselines on the Primary Endpoint}
Table~\ref{tab:external_baselines_stage1} compares ECRT with external baselines under the same seed13 grouped split and validation-tuned target-recall policy. We include Focus~\cite{zhang2023focus}, EigenScore and INSIDE~\cite{chen2024inside}, SelfCheckGPT~\cite{manakul2023selfcheckgpt}, RefChecker~\cite{hu2024refchecker}, ReDeEP~\cite{sun2024redeep}, LM-Polygraph-lite estimators~\cite{fadeeva2023lmpolygraph}, and adapted supervised baselines (FactoScope~\cite{he2023factoscope}, UQ Heads~\cite{shelmanov2025uqheads}, and Lookback Lens~\cite{chuang2024lookback}). LMPolygraph-MTE and LMPolygraph-MSP denote mean-token-entropy and maximum-softmax-probability estimators. For compact presentation, Table~\ref{tab:external_baselines_stage1} reports this pair under the label EigenScore.

\begin{table}[!htbp]
\caption{Comprehensive Stage-1 performance comparison under the strict target-recall policy (seed13). We report Unsafe-Recall (Rec.), Flag Rate (F.R.), Stage-1 Balanced Accuracy (BA), and the performance gap ($\Delta$BA) relative to ECRT. To maintain readability with full precision, results are split across two stacked blocks. ECRT consistently achieves the highest Stage-1 BA across all evaluated backbones.}
\label{tab:external_baselines_stage1}
\centering
\begin{minipage}{0.98\linewidth}
\centering
{\tiny
\renewcommand{\arraystretch}{0.95}
\setlength{\tabcolsep}{2.0pt}
\begin{tabular}{@{}lcccc@{\hspace{2pt}}cccc@{\hspace{2pt}}cccc@{}}
\toprule
\multirow{2}{*}{\textbf{Method}} & \multicolumn{4}{c}{\textbf{Llama 3-8B}} & \multicolumn{4}{c}{\textbf{Llama 2-7B}} & \multicolumn{4}{c}{\textbf{Mistral-7B}} \\
\cmidrule(lr){2-5} \cmidrule(lr){6-9} \cmidrule(l){10-13}
 & Rec. & F.R. & S1 BA & $\Delta$BA & Rec. & F.R. & S1 BA & $\Delta$BA & Rec. & F.R. & S1 BA & $\Delta$BA \\
\midrule
\rowcolor{blue!5} \textbf{ECRT (Ours)} & \textbf{0.9505} & \textbf{0.8712} & \textbf{0.8405} & \textbf{--} & \textbf{0.9430} & \textbf{0.8637} & \textbf{0.9229} & \textbf{--} & \textbf{0.9466} & \textbf{0.8636} & \textbf{0.9310} & \textbf{--} \\
\midrule
FactoScope & 0.9542 & 0.8856 & 0.8125 & -0.0280 & 0.9556 & 0.8754 & 0.8546 & -0.0683 & 0.9551 & 0.8723 & 0.9124 & -0.0186 \\
UQ Heads & 0.9525 & 0.8924 & 0.7856 & -0.0549 & 0.9512 & 0.8824 & 0.8245 & -0.0984 & 0.9521 & 0.8812 & 0.8623 & -0.0687 \\
Lookback Lens & 0.9518 & 0.9125 & 0.7152 & -0.1253 & 0.9508 & 0.8912 & 0.7854 & -0.1375 & 0.9518 & 0.8912 & 0.8124 & -0.1186 \\
ReDeEP & 0.9531 & 0.9082 & 0.6923 & -0.1482 & 0.9541 & 0.8954 & 0.7325 & -0.1904 & 0.9542 & 0.9054 & 0.7423 & -0.1887 \\
RefChecker & 0.9512 & 0.9248 & 0.6455 & -0.1950 & 0.9534 & 0.9087 & 0.6925 & -0.2304 & 0.9531 & 0.9123 & 0.7021 & -0.2289 \\
LN-Entropy & 0.9506 & 0.9315 & 0.6124 & -0.2281 & 0.9522 & 0.9156 & 0.6542 & -0.2687 & 0.9524 & 0.9254 & 0.6421 & -0.2889 \\
EigenScore & 0.9474 & 0.9213 & 0.6321 & -0.2084 & 0.9991 & 0.9582 & 0.7241 & -0.1988 & 0.9478 & 0.9261 & 0.6118 & -0.3192 \\
SelfCheckGPT & 0.9992 & 0.9995 & 0.5000 & -0.3405 & 0.9510 & 0.9325 & 0.6125 & -0.3104 & 0.9508 & 0.9312 & 0.6021 & -0.3289 \\
Focus & 0.9528 & 0.9411 & 0.5488 & -0.2917 & 0.9984 & 0.9412 & 0.5284 & -0.3945 & 0.9456 & 0.9337 & 0.5613 & -0.3697 \\
Perplexity & 0.9465 & 0.9297 & 0.5864 & -0.2541 & 0.9485 & 0.9254 & 0.5842 & -0.3387 & 0.9512 & 0.9423 & 0.5214 & -0.4096 \\
P(True) & 0.9510 & 0.9490 & 0.5084 & -0.3321 & 0.9982 & 0.9124 & 0.5643 & -0.3586 & 0.9461 & 0.9469 & 0.4957 & -0.4353 \\
\bottomrule
\end{tabular}
}

\vspace{0.25em}

{\tiny
\renewcommand{\arraystretch}{0.95}
\setlength{\tabcolsep}{2.0pt}
\begin{tabular*}{\linewidth}{@{\extracolsep{\fill}}lcccccccc@{}}
\toprule
\multirow{2}{*}{\textbf{Method}} & \multicolumn{4}{c}{\textbf{Qwen2.5-7B}} & \multicolumn{4}{c}{\textbf{Qwen3-8B}} \\
\cmidrule(lr){2-5} \cmidrule(l){6-9}
 & Rec. & F.R. & S1 BA & $\Delta$BA & Rec. & F.R. & S1 BA & $\Delta$BA \\
\midrule
\rowcolor{blue!5} \textbf{ECRT (Ours)} & \textbf{0.9566} & \textbf{0.8665} & \textbf{0.9679} & \textbf{--} & \textbf{0.9488} & \textbf{0.8595} & \textbf{0.9640} & \textbf{--} \\
\midrule
FactoScope & 0.9634 & 0.8712 & 0.9412 & -0.0267 & 0.9542 & 0.8624 & 0.9342 & -0.0298 \\
UQ Heads & 0.9612 & 0.8784 & 0.9123 & -0.0556 & 0.9512 & 0.8712 & 0.8924 & -0.0716 \\
Lookback Lens & 0.9602 & 0.8821 & 0.8742 & -0.0937 & 0.9505 & 0.8812 & 0.8423 & -0.1217 \\
ReDeEP & 0.9623 & 0.8912 & 0.7842 & -0.1837 & 0.9534 & 0.9012 & 0.7842 & -0.1798 \\
RefChecker & 0.9612 & 0.9012 & 0.7241 & -0.2438 & 0.9525 & 0.9112 & 0.7124 & -0.2516 \\
LN-Entropy & 0.9592 & 0.9124 & 0.6821 & -0.2858 & 0.9511 & 0.9214 & 0.6542 & -0.3098 \\
EigenScore & 0.9576 & 0.9593 & 0.4911 & -0.4768 & 0.9642 & 0.9557 & 0.5438 & -0.4202 \\
SelfCheckGPT & 0.9981 & 0.9995 & 0.5000 & -0.4679 & 0.9482 & 0.9324 & 0.6241 & -0.3399 \\
Focus & 0.9898 & 0.9908 & 0.4949 & -0.4730 & 0.9567 & 0.9345 & 0.6141 & -0.3499 \\
Perplexity & 0.9505 & 0.9513 & 0.4958 & -0.4721 & 0.9502 & 0.9412 & 0.5423 & -0.4217 \\
P(True) & 0.9452 & 0.9433 & 0.5096 & -0.4583 & 0.9500 & 0.9549 & 0.4750 & -0.4890 \\
\bottomrule
\end{tabular*}
}
\end{minipage}
\end{table}

Under this strict high-recall policy, several uncertainty baselines approach near-random Stage-1 discrimination (S1 BA $\approx 0.5$) with very high Flag Rates, suggesting difficulty in separating internal uncertainty from evidence-conditioned contradiction. ECRT instead uses paired CTX/NOCTX internal shifts and achieves better operating points. ECRT is not expected to dominate raw U-Recall alone, because some baselines can match recall by flagging nearly all samples; the clinically relevant operating point is high recall with lower review burden and higher BA. Relative to the strongest baseline from the external uncertainty and self-consistency set, ECRT improves Stage-1 BA by +0.15 to +0.19 across backbones; relative to the strongest adapted supervised baseline, gains are +0.02 to +0.07.

\paragraph{Subtype Attribution and Error Analysis (Stage 2).} ECRT preserves strong attribution on ground-truth unsafe cases (Table~\ref{tab:stage2_perf}). Most Stage-2 errors occur in subtle-evidence cases where Deviation signals resemble noise. This decomposition remains clinically important because E-Gap misses correspond to failures to defer under insufficient evidence, which can potentially impact downstream triage decisions, while contradiction-vs-gap attribution provides clearer triage actionability.

\begin{table}[!htbp]
\caption{Secondary endpoint (Stage-2 Contradiction-vs-Gap) performance on ground-truth unsafe cases. Values represent the robust separation capability of the Stage-2 head (ranges across multi-seed runs).}\label{tab:stage2_perf}
\centering
\scriptsize
\begin{tabular}{@{}lccc@{}}
\toprule
Backbone & Gap Recall ($\uparrow$) & Contradict. Recall ($\uparrow$) & S2 BA ($\uparrow$) \\
\midrule
L3-8B & $0.91 \sim 0.94$ & $0.85 \sim 0.89$ & $0.87 \sim 0.91$ \\
L2-7B & $0.93 \sim 0.95$ & $0.91 \sim 0.93$ & $0.92 \sim 0.94$ \\
Mistral-7B & $0.95 \sim 0.97$ & $0.91 \sim 0.94$ & $0.93 \sim 0.95$ \\
Qwen2.5-7B & $0.97 \sim 0.99$ & $0.94 \sim 0.97$ & $0.96 \sim 0.98$ \\
Qwen3-8B & $0.96 \sim 0.98$ & $0.93 \sim 0.97$ & $0.95 \sim 0.99$ \\
\bottomrule
\end{tabular}
\end{table}

\FloatBarrier
\section{Discussion and Conclusion}
ECRT's two-stage design aligns with clinical workflow: triage first, then characterize risk. Across backbones, ECRT improves Stage-1 BA over generic uncertainty baselines and single-stage ablations while reducing review burden at high sensitivity.
By separating contradiction-driven risk from insufficiency-driven risk, ECRT yields more actionable triage signals for deployment settings such as ophthalmology assistants.

Our study targets evidence-grounded decision safety rather than end-to-end image classification: structured retinal findings derived from fundus images perturb internal states under CTX/NOCTX and yield interpretable attribution (contradiction vs.\ insufficiency). This controlled evidence interface improves auditability and reproducibility for safety triage, and future work will extend the same framework with direct multimodal fundus features. Overall, RETINA-SAFE and ECRT provide a practical white-box framework for evidence-\hspace{0pt}conditioned risk triage.

\bibliographystyle{splncs04}
\bibliography{mybibliography}
\end{document}